\documentclass[letterpaper, 10 pt, conference]{ieeeconf}  % Comment this line out if you need a4paper

\IEEEoverridecommandlockouts                              % This command is only needed if 
\usepackage[caption=false,font=normalsize,labelfont=sf,textfont=sf]{subfig}
\usepackage{url}
\usepackage{verbatim}
\usepackage{amsmath,amssymb,amsfonts}
\usepackage{cite}
\usepackage{graphicx}
\usepackage{textcomp}
\usepackage{xcolor}
\usepackage{bbm}
\let\labelindent\relax
\usepackage{enumitem}
\usepackage{booktabs}
\usepackage{multirow}
\usepackage{soul}

\usepackage{times}
\usepackage{multicol}

\usepackage{graphicx}
\usepackage{bm}
\usepackage[nolist]{acronym}

\usepackage{algorithm}
\makeatletter
\let\For\relax
\let\EndFor\relax
\let\State\relax

\makeatother
\usepackage[noend]{algpseudocode}

\usepackage{mathtools}
\usepackage{array}
\usepackage{dblfloatfix}
\usepackage{diagbox}
\usepackage{etoolbox}
\AtBeginEnvironment{algorithmic}{\small}

\newif\ifshowcomments
\showcommentstrue  % to show
\ifshowcomments %jovin added
    \newcommand{\bae}[1]{\hl{[SB: #1]}\protect\color{black}} % Sangjae added, Aug 24, 2021 
    \newcommand{\di}[1]{\hl{[DI: #1]}\protect\color{black}} % David added, Aug 24, 2021
    \newcommand{\ft}[1]{\hl{[FT: #1]}\protect\color{black}} % Faizan
    \newcommand{\jd}[1]{\hl{[JD: #1]}\protect\color{black}} % Jovin
\else
    \newcommand{\bae}[1]{}
    \newcommand{\di}[1]{}
    \newcommand{\ft}[1]{}
    \newcommand{\jd}[1]{}
\fi

\usepackage{lipsum}
\usepackage[colorinlistoftodos]{todonotes}

\usepackage{amsthm}

\usepackage{bbm} % for mathbb 1
\usepackage[colorlinks = True, linkcolor = blue, citecolor = blue]{hyperref}
\DeclareMathOperator*{\argmin}{arg\,min}
\DeclareMathOperator*{\clip}{clip}
\DeclareMathOperator*{\wrap}{wrap}
\newcommand{\pmstd}[1]{\mathbin{\scriptstyle \pm\, #1}}
\usepackage{dsfont}  % for indicator func \mathds{1}
\newcommand{\method}{\textsf{\scalebox{0.95}[1.0]{HOLO-MPPI}}}
\newcommand{\scn}[1]{\texttt{\scalebox{0.8}[0.95]{#1}}}

\renewcommand{\baselinestretch}{0.99}

\title{
HOLO-MPPI: Multi-Scenario Motion Planning \\via Hierarchical Policy Optimization
}

\author{Youngjae Min$^{1*}$ \quad  Jovin D'sa$^2$ \quad Faizan M. Tariq$^2$ \quad David Isele$^2$  \quad Navid Azizan$^{1\dagger}$ \quad Sangjae Bae$^{2\dagger}$ 
\thanks{
$^*$The research work was done in part during an internship at HRI.
}
\thanks{
$^\dagger$Contributed equally to advising.
}
\thanks{$^1$Massachusetts Institute of Technology, Cambridge, MA 02139. (Email:\href{mailto:yjm@mit.edu}{\texttt{yjm@mit.edu}}, \href{mailto:azizan@mit.edu}{\texttt{azizan@mit.edu}})}
\thanks{$^2$Honda Research Institute, USA, San Jose, CA, 95134. (Email: \href{mailto:jovin_dsa@honda-ri.com}{\texttt{jovin\_dsa@honda-ri.com}}, \href{mailto:faizan_tariq@honda-ri.com}{\texttt{faizan\_tariq@honda-ri.com}}, \href{mailto:disele@honda-ri.com}{\texttt{disele@honda-ri.com}}, \href{mailto:sbae@honda-ri.com}{\texttt{sbae@honda-ri.com}}).}
}

\begin{document}

\maketitle
% \thispagestyle{empty}
% \pagestyle{empty}

%%%%%%%%%%%%%%%%%%%%%%%%%%%%%%%%%%%%%%%%%%%%%%%%%%%%%%%%%%%%%%%%%%%%%%%%%%%%%%%%

\begin{abstract}
Robots deployed in the real world must plan motions across diverse scenarios without per-scenario retuning. End-to-end reinforcement learning (RL) can generalize across scenarios but often becomes brittle under distribution shift, reward misspecification, and stochastic interactions. Model predictive path integral (MPPI) control enables strong real-time refinement without gradients, but its performance depends on a well-shaped sampling prior, while manually designing the priors does not scale to multi-scenario deployment. We present \method{} (High-level Offline, Low-level Online MPPI), a multi-scenario motion planning framework that combines high-level policy learning with low-level stochastic optimal control. Offline, we learn a high-level policy that proposes scenario-robust plans in an abstract action space, with a learned world model for online rollout. Online, the policy serves as a data-driven prior generator that parameterizes MPPI’s sampling distribution conditioned on the current observation and goal. MPPI then optimizes low-level control sequences around this prior in real time to adapt to local disturbances. We instantiate \method{} in autonomous driving by designing an effective high-level action space and tailored model architectures. Our evaluation across diverse driving scenarios shows that \method{} improves upon MPPI and end-to-end RL baselines while maintaining real-time control.
\end{abstract}

% \begin{IEEEkeywords}
% Motion planning, model predictive control, reinforcement learning, autonomous driving.
% \end{IEEEkeywords}

%%%%%%%%%%%%%%%%%%%%%%%%%%%%%%%%%%%%%%%%%%%%%%%%%%%%%%%%%%%%%%%%%%%%%%%%%%%%%%%%
\section{Introduction}

Robots deployed in the wild must plan motions across a spectrum of scenarios (e.g., varying layouts, dynamics, and task objectives) without per-scenario retuning. This multi-scenario setting is particularly challenging because the distribution of states, constraints, and disturbances can shift dramatically across environments, while planners must still produce safe, real-time behavior.

Reinforcement learning (RL) is an appealing approach to acquire policies directly from data, sidestepping the need to hand-engineer heuristics for each scenario. In principle, training on diverse environments can yield policies that generalize to unseen ones~\cite{cobbe2019quantifying}. In practice, however, end-to-end RL policies often exhibit unstable performance across scenarios due to long-horizon credit assignment, reward misspecification, and sensitivity to distribution shift and stochastic dynamics~\cite{hadfield2017inverse,dulac2021challenges}. These issues are exacerbated when learning directly in low-level action spaces (e.g., torques or fine-grained velocity commands), where policy-gradient variance is high~\cite{sutton1999between, vezhnevets2017feudal,nachum2018data}.

Model predictive path integral (MPPI) control, in contrast, is a sample-based optimizer for online policy refinement~\cite{williams2018information}. MPPI handles nonlinear dynamics without requiring gradients and exploits parallel sampling to meet real-time constraints. Its effectiveness, however, depends critically on the sampling prior, i.e., the distribution from which candidate control sequences are drawn~\cite{yin2022trajectory}. A well-shaped prior can dramatically improve sample efficiency and robustness, while a poor one wastes computation and degrades performance. Designing such priors by hand (e.g., scenario-specific trajectory libraries) does not scale to multi-scenario deployment.

\begin{figure}
    \centering
    \includegraphics[width=0.87\linewidth]{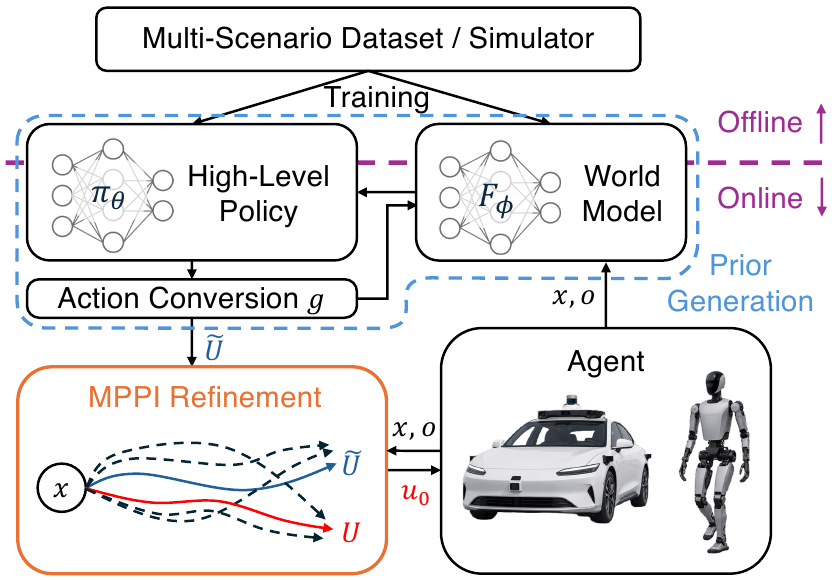}
    \caption{Schematic of \method{}. Offline, a high-level policy and a world model are trained on multi-scenario data. Online, the high-level policy and world model produce intention-level actions that are converted into a nominal low-level control sequence. MPPI samples around this prior, evaluates rollouts, and executes the first control of the optimized sequence.}
    \label{fig:schematic}
\end{figure}

% \IEEEpubidadjcol
This paper advocates a complementary decomposition: let learning supply scenario-robust, data-driven priors, and let online stochastic optimal control supply precise, constraint-aware refinement (see Figure~\ref{fig:schematic}). Moreover, we perform learning in a high-level action space that enables learning robust priors across multiple complex scenarios.

\textbf{Contributions}.
Our main contributions are as follows:
\begin{itemize}
    \item We introduce \method{}, a hierarchical, multi-scenario motion-planning framework that couples RL and MPPI through a learned sampling prior: the RL policy steers the search toward useful modes, while MPPI retains its exploration and constraint-aware refinement under the planner's full cost. This also decouples the RL reward from the MPC cost, letting each component use the objective it is best suited to optimize.
    \item 
    We learn the policy in an abstract action space designed to expose intent shared across scenarios, allowing a single planner to generalize across them without per-scenario retuning or handcrafted priors.
    \item 
    We instantiate \method{} in a multi-scenario highway-driving benchmark, jointly training the high-level policy with a lightweight world model for online rollout. The high-level policy is more scenario-robust than direct low-level RL, and \method{} further outperforms vanilla MPPI, end-to-end SAC, and the prior alone, with smoother control at real-time rates.
\end{itemize}

\section{Related Work}
\subsection{Sampling-based MPC and MPPI}
Sampling-based MPC methods such as MPPI and the cross-entropy method (CEM) have proven to be effective for nonlinear control under uncertainty~\cite{williams2018information, rubinstein1999cross, pinneri2021sample}.
MPPI updates a nominal control sequence using cost-weighted perturbations.
Its practical performance depends heavily on the sampling distribution used to generate rollouts.
Recent works improve MPPI via covariance shaping, adaptive control, and learned dynamics models~\cite{yin2022trajectory, mohamed2025toward,pravitra2020L,ryu2025iann}. Despite these advances, many approaches
still rely on heuristic or scenario-specific priors, which limit scalability to diverse multi-scenario settings.

\subsection{Learning-based priors and initializations for MPC}
A growing body of work integrates learning with MPC. Learning has been used to accelerate or initialize trajectory optimization by imitating MPC solutions~\cite{levine2013guided}, warm-starting MPC solvers~\cite{celestini2024transformer}, or learning high-level decision variables that parameterize MPC problems~\cite{song2022policy}.
Closer to our setting, recent work uses learned policies as MPPI sampling priors with learned value functions as terminal costs~\cite{hansen2022temporal, qu2023rl}.

\method{} differs in two key ways. First, it learns a high-level policy in an abstract action space designed to capture scenario-shared intent, enabling one planner to generalize across structurally distinct scenarios. Second, the learned policy is used only to shape the sampling distribution of MPPI, decoupling the RL reward from the MPC cost. 
Since RL typically needs reward shaping, an interpretable, constraint-aware MPC cost---often dictated by the problem---can be a poor RL reward; this decoupling lets each component use the objective it is best suited to optimize.

\subsection{Combining RL and MPC at deployments}

A separate line of work structurally combines RL and MPC at deployment. Some approaches place RL at the low level, training an RL policy to track or steer a model-based plan~\cite{jenelten2024dtc, zhang2026sumo}. Dual approaches place a learned high-level policy on top of a model-based low-level controller~\cite{yang2023cajun}, or blend RL with MPC via a learned torque residual~\cite{cheng2025rambo}.
Hierarchical decomposition has also been explored at the semantic level via vision-language models~\cite{schakkal2025hierarchical}.

\method{} most closely resembles the hierarchical pattern of high-level RL with low-level MPC, but differs in the high-level RL policy's role. Rather than producing a reference that the low-level controller deterministically tracks, the high-level policy parameterizes a sampling distribution over the low-level optimizer's solutions, allowing MPPI to explore around the prior under the planner's full cost.

% \subsection{Hierarchical and model-based reinforcement learning}
% Hierarchical RL decomposes decision-making into multiple
% levels of abstraction, such as options, skills, or subgoals, improving exploration
% and long-horizon credit assignment~\cite{sutton1999between, vezhnevets2017feudal,nachum2018data,schakkal2025hierarchical}.
% Model-based RL uses learned dynamics to plan or to generate imagined rollouts, improving sample efficiency compared to model-free approaches~\cite{chua2018deep,hafner2020dream}.

\section{Preliminaries}
\subsection{Problem setup: multi-scenario motion planning}

We consider a family of scenarios $\mathcal{S}$, where each scenario $s \in \mathcal{S}$
specifies the type of environment (e.g., merging or roundabout in autonomous driving). Our goal is to deploy a single planner that performs well across scenarios from $\mathcal{S}$.

At each initial state $x_0\in\mathcal{X}$ of the agent and the initial observation (or episode state) $o_0\in\mathcal{O}$, the planner solves a finite-horizon MPC
problem for the control sequence
$U:=u_{0:H-1}\!:=\!(u_0,u_1,\dots,u_{H-1})\in \mathcal{U}^H$ for planning horizon $H$:
\begin{equation}
\label{eq:problem}
\begin{aligned}
\argmin_{U \in \mathcal{U}^H} \;\;
& J_s(x_{0:H}, o_{0:H}, U)\!:=\!
\sum_{t=0}^{H-1}\! \ell_s(x_{t}, o_t, u_{t})
+ \Phi_s(x_{H}, o_H)\\
\text{s.t.}\;\;
& x_{t+1} = f(x_{t}, u_{t}),\\
& o_{t+1} = F_s(x_{t}, o_{t}, u_{t}),
\quad\quad t = 0, \dots, H\!-\!1,
\end{aligned}
\end{equation}
where $\ell_s$ is the stage cost, $\Phi_s$ is the terminal cost, $f$ is the dynamics of the agent, and $F_s$ is the (unknown) world dynamics for each scenario $s$. In a receding-horizon fashion, only the first action of the solution is executed before replanning at the next time step.

\subsection{Model Predictive Path Integral}
MPPI approximately solves the finite-horizon control problem in~\eqref{eq:problem} by evaluating control trajectories sampled around a nominal trajectory $\tilde{U}:=\tilde{u}_{0:H-1}\in \mathcal{U}^H$.

It considers an optimal stochastic policy (i.e., a distribution over $U$) for the following problem given covariance $\Sigma$~\cite{williams2018information}:
\begin{equation}
\label{eq:mppi_opt}
    q^*_\text{MPPI} = \argmin_{q} \mathbb{E}_{U\sim q}J_s(x_{0:H}, o_{0:H}, U)+\lambda \mathcal{D}_{KL}(q||q_{0,\Sigma}),
\end{equation}
which yields the analytical solution:
\begin{equation}
    q^*_\text{MPPI}(U) = \frac{1}{\eta_1}\exp\Big(-\frac{1}{\lambda}J_s(x_{0:H}, o_{0:H}, U)\Big) q_{0,\Sigma}(U),
\end{equation}
where $\eta_1$ is the normalization factor, and $q_{U,\Sigma}:=\mathcal{N}(\cdot|U,I_H\otimes\Sigma)$ denotes a multivariate Gaussian over the sequence of $H$ control inputs with mean $U$ and constant covariance $\Sigma$ at each timestep.
% Note that the KL-divergence becomes a quadratic control cost for $U$ if the optimization variable $q$ is restricted to the format $q_{U,\Sigma}$.

Since the optimal distribution $q^*_\text{MPPI}$ is not tractable, MPPI considers a controlled distribution $q_{U,\Sigma}$ and pushes it close to the optimal one by minimizing their distance:
\begin{equation}
    U^* = \argmin_U \mathcal{D}_{KL}(q^*_\text{MPPI}||q_{U,\Sigma})
    = \mathbb{E}_{U\sim q^*_\text{MPPI}}U.
\end{equation}
Then, it computes $U^*$ through importance sampling based on sampling distribution $q_{\tilde{U},\Sigma}$ around the nominal trajectory $\tilde{U}$:
\begin{equation}
    U^* = \mathbb{E}_{U\sim q^*_\text{MPPI}}U = \mathbb{E}_{U\sim q_{\tilde{U},\Sigma}} \frac{q^*_\text{MPPI}(U)}{q_{\tilde{U},\Sigma}(U)}U
    = \mathbb{E}_{U\sim q_{\tilde{U},\Sigma}} w(U) U,
\end{equation}
with weight:
\begin{equation}
\label{eq:mppi_weight}
    w(U) = \dfrac{1}{\eta_2}\exp\Big(-\frac{1}{\lambda}J_s(x_{0:H}, o_{0:H}, U) - \sum_{t=0}^{H-1}\tilde{u}_t^\top\Sigma^{-1}u_t\Big),
\end{equation}
where $\eta_2$ is the normalization factor. In practice, MPPI approximates the expectation by drawing $K$ samples $\{U^k\}_{k=1}^K$ from the sampling distribution $q_{\tilde{U},\Sigma}$:
\begin{equation}
\label{eq:mppi_update}
    U^* \simeq \sum_{k=1}^K w(U^k) U^k.
\end{equation}
In a receding-horizon fashion, MPPI executes the first action and uses the (zero-padded) remainder as the next prior $\tilde{U}$.

\section{Overview of \method{}}

\method{}, shown in Figure~\ref{fig:schematic}, is a hierarchical motion-planning framework for robots to operate across diverse scenarios without scenario-specific retuning. The central idea is to separate planning into two complementary layers:
\begin{itemize}
    \item \textbf{High-level offline policy learning}: we learn a policy $\pi_\theta$ via reinforcement learning that produces an abstract, scenario-robust plan that captures task intent in a compact high-level action space $\mathcal{U}_\text{high}$.
    \item \textbf{Low-level online policy optimization}: we refine the high-level plan online in the original low-level action space $\mathcal{U}$, enabling fast adaptation to local disturbances, model mismatch, and stochastic interactions.
\end{itemize}
This decomposition combines the strengths of learning and online optimization: RL provides a data-driven prior that generalizes across scenarios, while MPPI preserves the reactivity and constraint-awareness needed for real-time control.

The motivation for this design is that directly learning a policy that outputs low-level control actions is often difficult in multi-scenario settings. Low-level actions must account for fine-grained dynamics, long-horizon credit assignment, and highly variable local interactions, which can make training brittle and reduce transfer across environments. In contrast, a high-level action space can encode more stable and semantically meaningful decisions, such as subgoals, waypoints, motion primitives, mode switches, or other intention-level commands appropriate to the robotics domain. Learning in such an abstract space allows the policy to focus on decisions that are shared across scenarios, while deferring precise execution to the online optimizer.

Offline, \method{} learns both the high-level policy $\pi_\theta:\mathcal{X}\times\mathcal{O}\rightarrow\mathcal{U}_\text{high}$ and the world model $F_\phi:\mathcal{X}\times\mathcal{O}\times\mathcal{U}\rightarrow\mathcal{O}$ from data collected across scenarios from simulation, logged experience, exploration, or demonstrations. The world model predicts observations needed to roll out the high-level policy during prior generation, while the high-level policy maximizes long-horizon return through reinforcement learning. Importantly, \method{} does not require the high-level policy to be a perfect standalone controller. Its purpose is to generate a prior that places online optimization near useful solution modes across many scenarios.

At each planning step, given the initial state $x_0$ and observation $o_0$, we roll out the high-level actions from the learned policy $\pi_\theta$ and the learned model $F_\phi$. Then, we map the high-level actions into the original action space $\mathcal{U}$ through a known conversion function $g:\mathcal{X}\times\mathcal{O}\times\mathcal{U}_\text{high}\rightarrow\mathcal{U}$, which may correspond to a simple upsampling rule, a tracking controller for geometric references, or a structured controller that maps symbolic intent into dynamically feasible controls depending on the application. 

Given the converted rollout $\tilde{U}\in\mathcal{U}^H$, MPPI performs real-time stochastic optimization in the low-level action space. Specifically, it samples candidate control sequences around the learned prior, $U^k\sim\mathcal{N}(\tilde{U}, I_H\otimes\Sigma)$, evaluates the resulting rollouts under the task cost, and computes an importance-weighted update as in~\eqref{eq:mppi_update}. Crucially, the high-level policy steers MPPI toward promising regions of the control space, thereby improving sample efficiency and removing the need for handcrafted priors.

\begin{algorithm}[t]
    \caption{\method{}}\label{alg:method}
    \textbf{Input:} Initial state $x_0$; Initial observation $o_0$\\
    \textbf{Output:} Action sequence $U = (u_0, u_1, \cdots, u_{H-1})$.
    \newcommand{\AlgPhase}[1]{\Statex \vspace{0.2em}\textbf{#1}}
    \newcommand{\AlgStep}[1]{\Statex \vspace{0.1em}\hspace{-0.5em} \textcolor{gray}{#1}\vspace{0.1em}}
    \begin{algorithmic}[1]
        \Require Planning horizon $H$; Number of samples $K$; Sampling covariance matrix $\Sigma$; stage cost $\ell$; terminal cost $\Phi$; dynamics $f$; action-conversion model $g$; MPPI temperature scalar $\lambda$
        \AlgPhase{Phase 0: High-level offline policy learning}
        \State Learn high-level policy $\pi_\theta$ via reinforcement learning
        \State Learn world model $F_\phi$
        \AlgPhase{Phase 1: Low-level online policy optimization}
        % \While{not terminal} \Comment{Replan each control step}
        \AlgStep{(\textbf{1.1}) Roll out a nominal action prior using the high-level policy}
        \For{$t = 0, \dots, H \!-\! 1$}
            \State $u_\text{high} \leftarrow \pi_\theta(x_t, o_t)$
            \State $\tilde{u}_t \leftarrow g(x_t, o_t, u_\text{high})$
            \State $o_{t+1} \leftarrow F_\phi(x_t, o_t, \tilde{u}_{t})$
        \EndFor
        \AlgStep{(\textbf{1.2}) Sample and evaluate trajectories around the prior in parallel}
        \For{$k = 1, \dots, K$}
            \State Sample a control sequence $U^k = u^k_{0:H-1}\sim \mathcal{N}(\tilde{U}, I_H\otimes\Sigma)$
            \State $J^k \leftarrow 0$
            \For{$t = 0, \dots, H \!-\! 1$}
                \State $x_{t+1} \leftarrow f(x_{t}, u^k_{t})$
                \State $o_{t+1} \leftarrow F_\phi(x_{t}, o_t, u^k_{t})$
                \State $J^k \leftarrow J^k + \ell(x_t, o_t, u^k_{t}) + \lambda \tilde{u}^T_{t}\Sigma^{-1}u^k_t$
            \EndFor
            \State $J^k \leftarrow J^k + \Phi(x_H, o_{H})$
        \EndFor
        \AlgStep{(\textbf{1.3}) Compute importance weights and update the action sequence}
        \State $\beta \leftarrow \min_k J^k$
        \State $\eta \leftarrow \Sigma_{k=1}^{K}\exp\left(  -\frac{1}{\lambda}\left(J^k- \beta \right)  \right)$ 
        \For{$k = 1, \dots, K$}
            \State $\omega^k \leftarrow \frac{1}{\eta}  \exp \left( -\frac{1}{\lambda}\left(J^k- \beta \right)  \right)$
        \EndFor
        \State \Return $\sum_{k=1}^{K} \omega^k U^k$
    % \EndWhile
    \end{algorithmic}
\end{algorithm}

Algorithm~\ref{alg:method} summarizes the resulting procedure. At each control cycle, \method{} first queries the high-level policy using the current observation and goal, then converts the predicted abstract plan into a nominal low-level action sequence $\tilde{U}$, and finally runs MPPI around this learned prior to produce the control that is executed on the robot.

\method{} is a general framework: the same architecture admits different modeling choices depending on the robotic platform and available prior knowledge. In particular, it is compatible with both on-policy RL in simulation and off-policy RL from logged data. For rollout-based evaluation, it can incorporate analytic models, learned models, or hybrid combinations of the two. When accurate dynamics are available, the ego system can use the known model with analytic or learned predictions for the surroundings.

In the next section, we describe \method{} in more detail, focusing on the prominent application of autonomous driving.
We instantiate its task-dependent ingredients, such as the MPC formulation in~\eqref{eq:problem}, the design of the high-level action model with its corresponding action conversion mechanism, and the training process of the high-level policy and the world model.

\section{\method{} in Autonomous Driving}
\label{sec:holo_ad}

\begin{figure}[t]
    \centering
    \includegraphics[width=.95\linewidth]{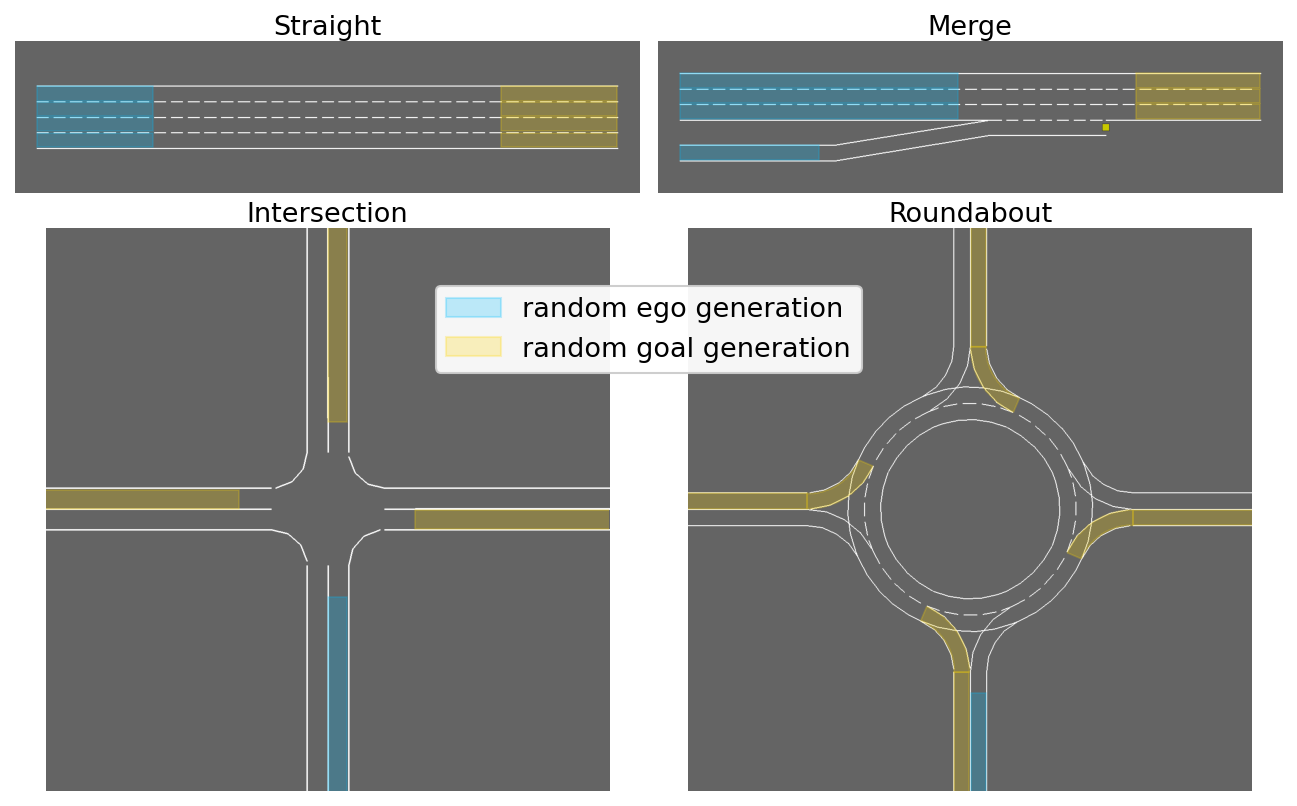}
    \caption{Four highway driving scenarios: \scn{Straight}, \scn{Merge}, \scn{Roundabout}, and \scn{Intersection}. The initial position of the ego vehicle and the goal position are randomly generated in the marked area. We randomize lane numbers, spawn/goal locations, and traffic behaviors.}
    \label{fig:scenarios}
\end{figure}

We instantiate \method{} in a multi-scenario highway-driving benchmark that we built on the Highway-Env simulator~\cite{highway-env}. Each episode samples one scenario from \scn{Straight}, \scn{Merge}, \scn{Roundabout}, and \scn{Intersection} (see Figure~\ref{fig:scenarios}), which exposes a common observation interface to both the learned policy and the online planner. 
In addition to vehicle states, BEV images, and the ego-frame goal, the planner has access to the local lane graph in the ego neighborhood, including lane center-lines and connectivity. This local map information is treated as part of the driving interface, as in standard map or perception-based autonomous driving stacks. It does not provide scenario labels, expert trajectories, or scenario specific tuning.
The observation $o\in\mathcal{O}$ contains (i) kinematic states $o^\text{kin}\in\mathbb{R}^{10\times7}$ for up to ten vehicles, with per-vehicle features \([{\rm presence}, x, y, v_x, v_y, \cos\psi, \sin\psi]\), (ii) an \(80\times 80\) grayscale bird's-eye-view image $o^\text{img}\in\mathbb{R}^{80\times80}$, and (iii) the goal position $o^\text{goal}\in\mathbb{R}^{2}$ expressed in the ego frame.
The online planner operates on low-level actions $u_t=[a_t,\delta_t]$, where $a_t$ is longitudinal acceleration and $\delta_t$ is steering angle.

\subsection{MPC formulation}

At test time, MPPI optimizes a horizon-$H$ sequence of low-level controls around a learned prior. The state of the ego vehicle is propagated with a kinematic bicycle model with control period $\Delta t = 0.2s$. Writing the ego state as $x_t = [p_{x,t}, p_{y,t}, \psi_t, v_t]$, 
the dynamics $f$ is represented as
\begin{equation}
f(x_t)=
\begin{bmatrix}
p_{x,t+1}\\
p_{y,t+1}\\
\psi_{t+1}\\
v_{t+1}
\end{bmatrix}
=
\begin{bmatrix}
p_{x,t}\\
p_{y,t}\\
\psi_t\\
v_t
\end{bmatrix}
+
\begin{bmatrix}
v_t \cos(\psi_t+\beta_t)\\
v_t \sin(\psi_t+\beta_t)\\
\dfrac{v_t \sin\beta_t}{l}\\
a_t
\end{bmatrix}\Delta t,
\end{equation}
where $\beta_t = \arctan\!\left(\frac{\tan \delta_t}{2}\right)$ is the slip angle, and $l$ is the half length of the ego vehicle used by the simulator.

For the objective optimized by the planner, we use the same cost across all scenarios:
\begin{align}
\ell(x_t,o_t,u_t) &=
w_v (v_t - v^\star)^2
+ w_a a_t^2
+ w_\delta \delta_t^2
+ w_{\rm lat} d_{\rm lat}(x_t)^2 \nonumber \\
&\quad + w_{\rm off}\mathds{1}\!\left[{\rm offroad}(x_t)\right]
+ w_{\rm col}\mathds{1}\!\left[{\rm collision}(x_t)\right],
\end{align}
where $v^\star=9$ m/s is the desired cruising speed and $d_{\rm lat}(x_t)$ is the lateral offset from the closest valid lane centerline. A rollout is terminated early when the ego reaches the goal, collides, or leaves the road; in the code, the arrival threshold is 1.5 m from the goal. The terminal cost is configurable and includes the options
\begin{equation} \label{eq:terminal_cost}
\Phi(x_H,o_H)\in\left\{w_{\rm term}\|g_H\|_2,\ w_{\rm term}\frac{100\|g_H\|_2}{10+\|g_H\|_2}\right\},
\end{equation}
where $g_H$ is the goal vector in the ego frame. In our experiments, we use $w_v=w_a=w_\delta=w_{\rm lat}=0.1, w_{\rm off}=w_{\rm col}=100$, and $w_{\rm term}\in\{1,5,10\}$.
For MPPI hyperparameters, we use $H=8$ with $K=1000$ samples, sampling covariance $\Sigma=\text{diag}(0.8^2,0.157^2)$ and temperature $\lambda=0.01$.

\subsection{High-level action model} \label{sec:AD_action}
We construct a compact high-level action space $\mathcal{U}_\text{high}$ by adapting the simulator's discrete meta-action model~\cite{highway-env} to preserve the Markov property needed for reinforcement learning. At each decision step, the policy selects one semantic driving command from $\mathcal{U}_\text{high}=\{\scn{IDLE},\scn{FASTER},\scn{SLOWER},\scn{LANE\_LEFT},\scn{LANE\_RIGHT}\}$. Each high-level action $h\in\mathcal{U}_\text{high}$ is then converted into a low‑level acceleration and steering action,
\begin{equation}
g(x,o,h)
=
\begin{bmatrix}
a(x,h)\\
\delta(x,o,h)
\end{bmatrix},
\end{equation}
for the ego state $x=[p_x,p_y,\psi,v]\in\mathcal{X}$ and observation $o\in\mathcal{O}$.

For longitudinal control, the current speed $v$ is first mapped to the nearest target speed in a discrete grid $\mathcal{V}=(v^i)_{i=1}^M=(0,4.5,9)$. The $\scn{FASTER}$ and $\scn{SLOWER}$ actions increase or decrease the target-speed index by one, while all other commands keep the current target-speed index unchanged:
\begin{equation}
i^*(v,h)= \clip_{[1,M]} \left(
\argmin_{i\in[1,M]} |v^i-v| + \Delta (h)\right),
\end{equation}
where $\Delta (h) := \mathds{1}[h=\scn{FASTER}] - \mathds{1}[h=\scn{SLOWER}]$ is the index shift, and $\clip_{[1,M]} z := \min\left(M,\max(0,z)\right)$.
The resulting target speed $v^*(v,h)=v^{i^*(v,h)}$ is then tracked by a proportional controller:
\begin{equation}
a(x,h)=
\clip_{[-a_{\max},a_{\max}]}
K_a\big(v^*(v,h)-v\big),
\end{equation}
where \(K_a=1/0.6\) and \(a_{\max}=5\ {\rm m/s^2}\).

For lateral control, $\scn{LEFT\_LANE}$ and $\scn{RIGHT\_LANE}$ select the closest feasible adjacent lane in the commanded direction using the local lane graph available to the planner. This conversion utilizes the local lane graph introduced above without any scenario-specific controller parameters or maneuver libraries. The lane selection respects lane boundaries and road connectivity. If no feasible adjacent lane exists, the target lane remains the current lane. Non-lane-change actions also keep the current lane as the target.

The target lane is tracked with a cascaded lateral controller for smooth and bounded steering commands. Let $(s_\ell(p),d_\ell(p))$ be the local longitudinal and lateral coordinates of the ego position $p=[p_x,p_y]$ in lane $\ell$, and let $\theta_\ell(s)$ be the lane heading at longitudinal coordinate $s$. For the selected target lane $\ell_\text{tar}$, we use a short lookahead heading 
$
\theta_{\rm fut}
=
\theta_{\ell_{\rm tar}}\bigl(s_{\ell_\text{tar}}(p)+v\tau_p\bigr)$
with $\tau_p=0.1$s. The controller then converts lane-tracking error into steering through three steps. First, the lateral offset to the target lane is converted into a desired lateral velocity $v_{\rm lat}=-K_\text{lat} d_{\ell_\text{tar}}(p)$ with $K_\text{lat}=0.5$. Then, this velocity is converted into a bounded heading correction relative to the lookahead lane heading:
\begin{equation}
    \psi_\text{ref} =
    \theta_\text{fut}
    +
    \clip_{[-\bar{\psi},\bar{\psi}]}
    \sin^{-1}\left(
    \clip_{[-1,1]}
    \frac{v_{\rm lat}}{\operatorname{nz}(v)}
    \right),
\end{equation}
where $\bar{\psi}=\pi/4$, and
$\operatorname{nz}(v):=\operatorname{sign}(v)\max(|v|,10^{-2})$ avoids division by zero.
Third, the resulting heading error is converted into a desired yaw rate and then into a steering command through the bicycle-model geometry:
\begin{align*}
&\dot{\psi}_{\rm cmd}
=
K_\psi\,\wrap_{[-\pi,\pi]}(\psi_{\rm ref}-\psi),
\gamma
=
\sin^{-1}\!\!\left(
\clip_{[-1,1]}
\frac{l}{\operatorname{nz}(v)}\dot{\psi}_{\rm cmd}
\right),\\
&\delta(x,o,h)
=
\clip_{[-\delta_{\max},\delta_{\max}]}
\tan^{-1}\left(2\tan\left(\clip_{[-\pi/2+\epsilon, \pi/2-\epsilon]} \gamma\right)\right),
\end{align*}
where $K_\psi=3.0, \delta_{\max}=\pi/3$, and $\epsilon=10^{-3}$ prevents numerical singularities of evaluating $\tan(\cdot)$ at $\pm \pi/2$.

This action model exposes a small set of interpretable driving intentions to the high-level policy while guaranteeing bounded, dynamically meaningful low-level controls. Because the target speed and target lane are deterministic functions of the current state, observation, action, and local lane graph, the resulting high-level decision process remains Markov with respect to the planner’s available information.

\subsection{Reinforcement learning of high-level policy}
We train the high-level policy with DQN across all driving scenarios. We use a dense reward that encourages fast goal reaching while penalizing accidents (collisions or off-road events), lane deviation, and control effort:
\begin{equation*}
    r = 10\mathds{1}\{\text{arriv}\} -5\mathds{1}\{\text{accid}\} + 0.02r_\text{speed} -0.01( r_\text{lane} + r_\text{control})
\end{equation*}
where $\mathds{1}\{\cdot\}$ is the indicator function for an event, and $r_\text{speed} = \text{clip}\big(\frac{v-4.5}{4.5},-1,1\big), r_\text{lane} = \min\big(\frac{d_\text{lane}^2}{5},1\big), r_\text{control} = \big[(\frac{a}{5})^2+(\frac{\delta}{\pi/3})^2\big]$.
Episodes terminate on goal reaching, collision, off-road departure, or timeout.

\begin{figure}[t]
    \centering
    \includegraphics[width=.85\linewidth]{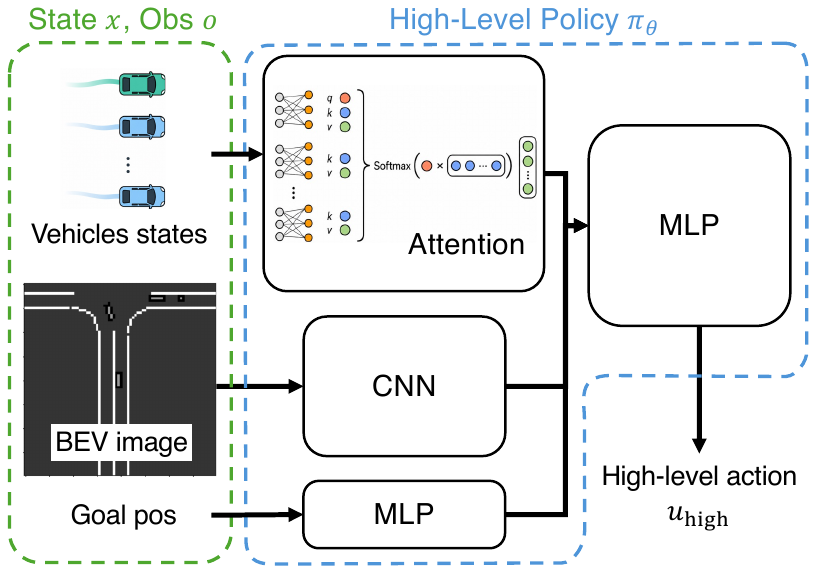}
    \caption{Architecture of the high-level policy model. A permutation-invariant attention-based vehicle-state encoder, a BEV image encoder, and a goal encoder are fused to generate a high-level driving intention that can be mapped to a nominal low-level action.}
    \label{fig:model}
\end{figure}

The policy network follows the architecture in Fig.~\ref{fig:model}. We utilize an ego-centric attention module as in~\cite{leurent2019social} to reflect the permutation-invariant structure of the vehicles' states. Separate embeddings are applied for the ego and neighboring vehicles, followed by two-head attention over the set of observed vehicles. The BEV image is encoded by a three-layer convolutional network, and the goal vector is encoded by a small multilayer perceptron. These three embeddings are concatenated and passed to the MLP head. In the configuration used in our experiments, the kinematic, image, and goal branches output 64, 64, and 16 features, respectively, followed by a \([128,64]\) multilayer head. We train with 40 parallel environments for 7M environment steps, replay-buffer size 0.5M, batch size 1024, learning rate \(3\times 10^{-4}\), discount factor \(0.99\), and an \(\epsilon\)-greedy schedule with exploration fraction \(0.2\) and final \(\epsilon=0.05\).

\subsection{World model}

Our online rollout model is hybrid rather than fully learned. Ego kinematics are propagated analytically with the bicycle model above, while neighboring vehicles follow a constant-velocity model in the world frame for simplicity and speed. Learning the behavior of neighboring vehicles as part of the world model is a natural extension for more accurate prediction in complex settings such as urban driving, where interactive and less structured behaviors are more prevalent.

We additionally train a learned predictor for the next BEV image. Given the current observation and low-level action, the current image is first shifted through a map $T$ according to the ego translation, and the network $F_\phi^\text{img}$ predicts only the residual correction:
\begin{equation}
o^\text{img}_{t+1} = T(x_t, o_t, u_t) + F_\phi^\text{img}(x_t, o_t, u_t).
\end{equation}
The predictor uses the same kinematic, image, and goal encoders as the policy, adds an action encoder, and decodes the fused latent through a skip-connected deconvolutional network. Training data are drawn directly from the replay buffer used for RL, augmented with the executed low-level actions. The loss is a weighted pixelwise reconstruction objective over multi-step rollouts:
\begin{equation}
\mathcal{L}_{\rm img} =
\frac{\sum_{t,p} w_{t,p}\,\|o^\text{img}_t(p)-o^\text{true}_t(p)\|_1}
{\sum_{t,p} w_{t,p}},
\end{equation}
where the per-pixel weights increase on foreground and edge:
\begin{equation*}
w_{t,p}=1+\lambda_{\rm fg}\mathds{1}[o^\text{true}_t(p)>\tau_{\rm fg}]
+\lambda_{\rm edge}\mathds{1}[\|\nabla o^\text{true}_t(p)\|>\tau_{\rm edge}].
\end{equation*}
In our implementation, \(\lambda_{\rm fg}=4.0\), \(\lambda_{\rm edge}=2.0\), \(\tau_{\rm fg}=0.55\), and \(\tau_{\rm edge}=0.08\). We train with Adam with a learning rate \(3\times 10^{-4}\), batch size 512, four-step rollout loss, and a teacher-forcing ratio linearly decayed from \(1.0\) to \(0.25\).

At planning time, image prediction is only needed for generating the nominal action prior. Once the prior is generated, the planner rolls out only the kinematic branches analytically and skips image prediction entirely for evaluating the sample control sequences around the prior.

\subsection{Collision evaluation}
We check collision by testing pairwise overlap between the ego vehicle and each neighboring vehicle using the separating axis theorem (SAT)~\cite{gottschalk1996obbtree}. SAT guarantees that two convex polygons do not intersect if there exists an axis along which their projections are disjoint. Vehicles are modeled as oriented rectangles. For each ego–neighbor pair, the algorithm evaluates the ego longitudinal and lateral axes together with the neighbor longitudinal and lateral axes.
% The relative center displacement is projected onto each axis and compared to the sum of the projected half‑extents on that axis. If any axis shows a gap larger than this bound (with a small tolerance), the rectangles are separated and no collision is reported; otherwise, they intersect.
% This SAT‑based test provides a geometry‑consistent collision indicator, computed efficiently in batch form and integrated directly into the cost function.

Let $T\in\mathbb{R}^2$ denote the center displacement and let $v_1,w_1,v_2,w_2\in\mathbb{R}^2$ be the two body-fixed axes of the ego and neighbor vehicles. Each ego–neighbor pair is declared separated if any of the four projections satisfy
\begin{equation*}
\exists a\in\{v_1,w_1,v_2,w_2\} \text{ s.t. } |T^\top a| > r_\text{ego}(a)+r_\text{neighbor}(a),
\end{equation*}
where $r_{(\cdot)}(a)$ is the projected half-extent on axis $a$. Otherwise, the pair intersects. We evaluate all MPPI samples and neighbors in parallel, enabling real-time planning.

\section{Experiments}

\subsection{Experimental setup}

We evaluate \method{} on the benchmark introduced in Section~\ref{sec:holo_ad} across all four scenarios. Each method is tested over 400 episodes (100 per scenario)

We compare \method{} against the following baselines.
\begin{itemize}
    \item \textbf{SAC}: RL policy trained via SAC acting directly in the original low-level action space $\mathcal{U}$.
    \item \textbf{Prior}: Our high-level policy trained via DQN that is used as a sampling-prior generator for \method{}.
    \item \textbf{FHP}: Filtered high-level policy instances. Since the high-level policy is trained offline, collapsed training runs can be identified and excluded before deployment.
    \item \textbf{MPPI}: Generic MPPI employing the previous plan as a prior (the initial prior is zero).
    \item \textbf{SAC/FHP \!+\! MPPI}: MPPI with a prior generated by SAC or FHP as an ablation study.
\end{itemize}

We report the success rate (goal reached without collisions or going off-road) along with per-step control effort and comfort, defined as the squared sum of control action and its first-order difference, respectively.

\subsection{Reinforcement learning of high-level priors}

\begin{figure}[t]
    \centering
    \includegraphics[width=\linewidth]{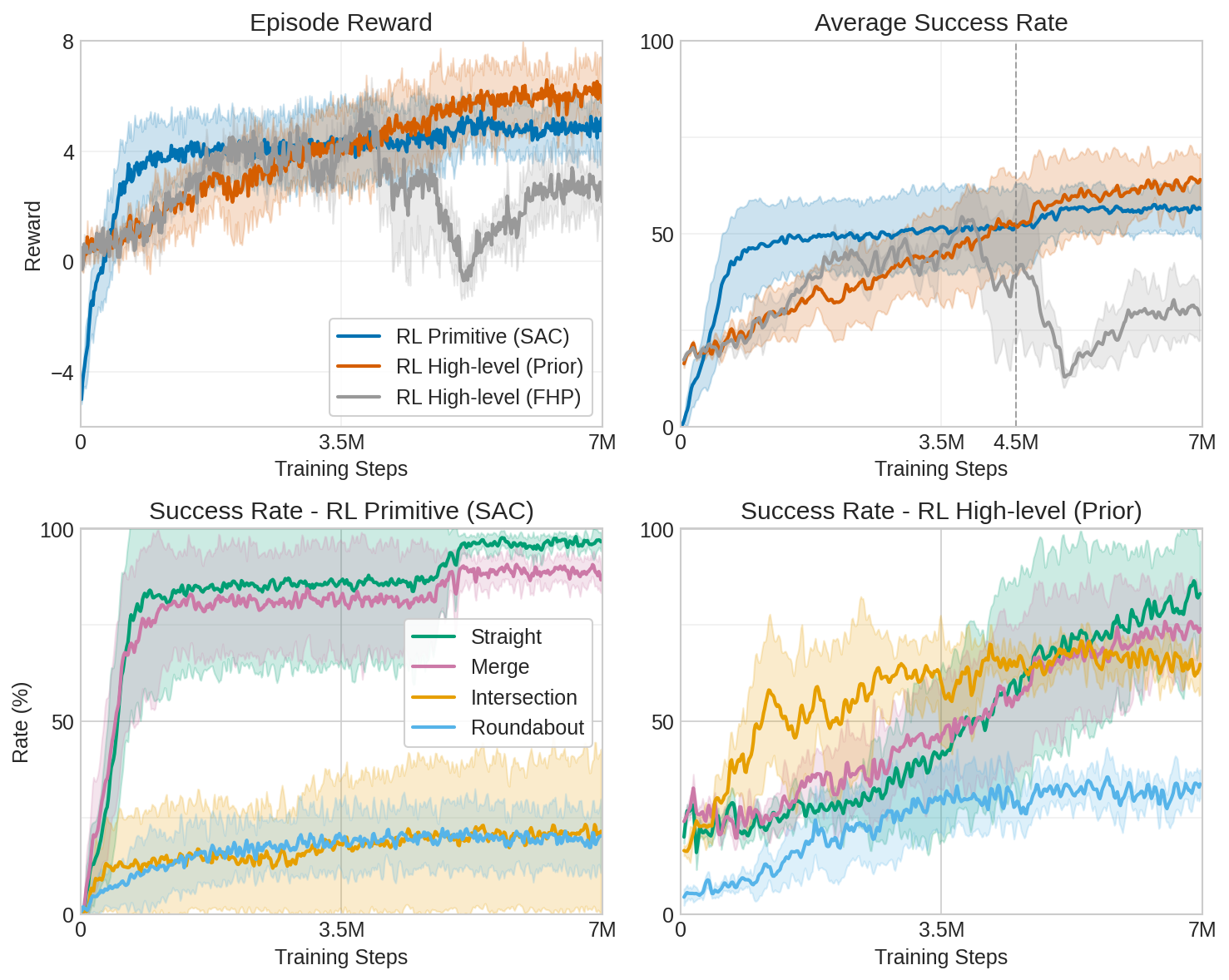}
    \caption{Learning curves over 10 training runs for low-level (SAC) and the high-level policy (Prior). The top row shows episode reward and average success rate over training, while the bottom row breaks down success rate by scenario. Among the 10 high-level-policy runs, 3 collapsed during offline training and are reported separately as FHP.}
    \label{fig:learning_curves}
\end{figure}

Figure~\ref{fig:learning_curves} compares the learning dynamics of the low-level policy (SAC) and the high-level policy (Prior) that is later used to parameterize MPPI's sampling distribution. 
SAC improves more rapidly during the early stage of training, suggesting that it quickly exploits the easier modes of the scenario distribution. However, the per-scenario curves on the lower left plot show that its performance is highly uneven: SAC achieves strong results on the simpler (\scn{Straight} and \scn{Merge}) scenarios, but remains much weaker on the more complex (\scn{Intersection} and \scn{Roundabout}) scenarios. In contrast, the high-level policy improves more gradually, yet ultimately attains a higher overall reward and average success rate while exhibiting a substantially more balanced performance profile across scenarios.

This distinction is important for \method{}. The learned policy is not the final controller, but it is used as a sampling-prior generator for the downstream MPPI planner. In this role, broad competence across the scenarios is more valuable than near-optimal performance on only a subset of tasks. A prior that is consistently reasonable across scenarios provides a better warm start for MPPI and reduces the likelihood that online optimization begins from a poor mode. We therefore use our designed high-level policy as the nominal prior generator in the downstream \method{} experiments.

\subsection{Main results}

\begin{table}[t]
\centering
\caption{Performance evaluated over 400 episodes (100 per scenario) with mean $\pm$ std over 10 trained RL models. \method{} achieves the highest overall success rate while maintaining smoother control.}
\label{tab:planner_single}
\setlength{\tabcolsep}{2pt}
\begin{tabular}{@{}c|cccc}
\toprule
\textbf{Scenario} & \textbf{Method} & \textbf{Success (\%) $\uparrow$} & \textbf{Ctrl Effort $\downarrow$} & \textbf{Ctrl Comfort $\downarrow$} \\
\midrule
\multirow{5}{*}{Overall} & MPPI & 57.8 & {\boldmath $0.43 \pmstd{0.27}$} & {\boldmath $0.17 \pmstd{0.08}$} \\
& SAC & $56.2 \pmstd{7.6}$ & $9.04 \pmstd{3.89}$ & $8.67 \pmstd{6.02}$ \\
& Prior & $64.9 \pmstd{8.0}$  & $1.64 \pmstd{2.83}$ & $3.31 \pmstd{6.22}$ \\
& SAC \!+\! MPPI & $56.6 \pmstd{6.9}$ & $7.98 \pmstd{2.62}$ & $8.42 \pmstd{4.02}$ \\
& \method{} & {\boldmath $68.0 \pmstd{7.2}$} & $1.27 \pmstd{1.94}$ & $2.20 \pmstd{3.74}$ \\
\hline\rule{0pt}{2.ex}
\multirow{5}{*}{Straight} & MPPI & 85.0 & {\boldmath $0.33 \pmstd{0.10}$} & {\boldmath $0.14 \pmstd{0.04}$} \\
& SAC & {\boldmath $97.5 \pmstd{2.5}$} & $7.96 \pmstd{3.75}$ & $7.79 \pmstd{6.38}$ \\
& Prior & $77.3 \pmstd{17.0}$  & $1.40 \pmstd{2.95}$ & $3.36 \pmstd{7.63}$ \\
& SAC \!+\! MPPI & $96.6 \pmstd{2.3}$ & $7.46 \pmstd{2.53}$ & $8.45 \pmstd{4.26}$ \\
& \method{} & $78.9 \pmstd{14.8}$ & $0.98 \pmstd{1.79}$ & $2.02 \pmstd{4.27}$ \\
\hline\rule{0pt}{2.ex}
\multirow{5}{*}{Merge} & MPPI & 59.0 & {\boldmath $0.44 \pmstd{0.19}$} & {\boldmath $0.17 \pmstd{0.05}$} \\
& SAC & $87.9 \pmstd{3.9}$ & $9.19 \pmstd{3.36}$ & $8.91 \pmstd{5.91}$ \\
& Prior & $73.7 \pmstd{12.6}$  & $1.38 \pmstd{2.31}$ & $3.06 \pmstd{5.44}$ \\
& SAC \!+\! MPPI & {\boldmath $89.6 \pmstd{4.2}$} & $8.02 \pmstd{2.31}$ & $8.45 \pmstd{4.07}$ \\
& \method{} & $74.1 \pmstd{13.5}$ & $1.15 \pmstd{1.57}$ & $2.11 \pmstd{3.20}$ \\
\hline\rule{0pt}{2.ex}
\multirow{5}{*}{Intersection} & MPPI & 31.0 & {\boldmath $0.45 \pmstd{0.15}$} & {\boldmath $0.18 \pmstd{0.05}$} \\
& SAC & $22.9 \pmstd{21.9}$ & $10.21 \pmstd{3.22}$ & $9.65 \pmstd{3.69}$ \\
& Prior & {\boldmath $68.9 \pmstd{6.5}$}  & $1.99 \pmstd{3.17}$ & $3.39 \pmstd{5.78}$ \\
& SAC \!+\! MPPI & $22.9 \pmstd{21.3}$ & $8.39 \pmstd{2.33}$ & $8.10 \pmstd{2.81}$ \\
& \method{} & $68.3 \pmstd{6.4}$ & $1.58 \pmstd{2.31}$ & $2.44 \pmstd{3.89}$ \\
\hline\rule{0pt}{2.ex}
\multirow{5}{*}{Roundabout} & MPPI & {\boldmath $56.0$} & {\boldmath $0.57 \pmstd{0.45}$} & {\boldmath $0.21 \pmstd{0.13}$} \\
& SAC & $16.6 \pmstd{7.8}$ & $13.04 \pmstd{4.83}$ & $11.25 \pmstd{5.92}$ \\
& Prior & $39.6 \pmstd{5.5}$  & $1.97 \pmstd{2.77}$ & $3.53 \pmstd{5.11}$ \\
& SAC \!+\! MPPI & $17.1 \pmstd{6.1}$ & $10.13 \pmstd{3.62}$ & $8.49 \pmstd{3.81}$ \\
& \method{} & $50.7 \pmstd{7.4}$ & $1.49 \pmstd{2.04}$ & $2.28 \pmstd{3.37}$ \\
\bottomrule
\end{tabular}
\end{table}

\begin{table}[t]
\centering
\caption{Success rate (\%) evaluated over 400 episodes (100 per scenario) with mean $\pm$ std over 10 trained RL models from three different train steps. For the methods with MPPI refinements, we report the mean and best performances across the six terminal cost settings in~\eqref{eq:terminal_cost}.}
\label{tab:planner_multi}
\setlength{\tabcolsep}{5pt}
\begin{tabular}{l|ccc}
\toprule
\multirow{2}{*}{\textbf{Method}}
& \multicolumn{3}{c}{\textbf{Train Step}} \\
\cline{2-4}
\rule{0pt}{2.5ex} & \textbf{3.5M}
& \textbf{4.5M}
& \textbf{7M} \\
\midrule
MPPI (mean)
% & \multicolumn{3}{c}{$44.3$}\\
& $44.3$ & $44.3$ & $44.3$\\
MPPI (best)
% & \multicolumn{3}{c}{{\boldmath $58.0$}}\\
& {\boldmath $58.0$} & $58.0$ & $58.0$\\
\rule{0pt}{2.5ex}SAC
& $50.1 \pmstd{10.9}$
& $50.6 \pmstd{11.2}$
& $56.2 \pmstd{7.6}$ \\
SAC \!+\! MPPI (mean)
& $51.1 \pmstd{10.6}$
& $51.9 \pmstd{10.5}$
& $56.9 \pmstd{7.4}$ \\
SAC \!+\! MPPI (best)
& $52.2 \pmstd{10.8}$
& $53.2 \pmstd{10.7}$
& $57.9 \pmstd{7.2}$ \\
\rule{0pt}{2.5ex}FHP
& $48.9 \pmstd{7.4}$
& $43.2 \pmstd{15.9}$
& $31.5 \pmstd{9.5}$ \\
FHP \!+\! MPPI (mean)
& $55.3 \pmstd{6.5}$
& $49.8 \pmstd{17.0}$
& $41.8 \pmstd{8.8}$ \\
FHP \!+\! MPPI (best)
& $56.9 \pmstd{7.4}$
& $52.0 \pmstd{17.4}$
& $43.5 \pmstd{9.1}$ \\
\rule{0pt}{2.5ex}Prior
& $42.4 \pmstd{15.9}$
& $53.3 \pmstd{12.6}$
& $64.9 \pmstd{8.0}$ \\
\method{} (mean)
& \shortstack{$49.8 \pmstd{13.7}$}
& \shortstack{$57.1 \pmstd{13.2}$}
& \shortstack{$67.3 \pmstd{6.9}$} \\
\method{} (best)
& \shortstack{$51.6 \pmstd{13.6}$}
& \shortstack{{\boldmath $58.6 \pmstd{13.3}$}}
& \shortstack{{\boldmath $69.1 \pmstd{6.5}$}} \\
\bottomrule
\end{tabular}
\end{table}

Table~\ref{tab:planner_single} summarizes performance across the four scenarios. Overall, \method{} achieves the highest success rate while maintaining competitive control smoothness. Vanilla MPPI attains the lowest control effort, but this is unsurprising: without a learned prior, it tends to produce conservative, low-magnitude actions that fail to make progress when more aggressive motion is required, as seen in the \scn{Intersection} scenario where turning maneuvers are necessary. The learned high-level policy alone (Prior) is already competitive, confirming that the abstract action space captures useful scenario-level intent. Coupling it with online MPPI further improves both success rate and control smoothness. We note that although these gaps appear to fall within the error bounds, the comparison is paired: MPPI is applied to the same models on the same episodes, so the gaps reflect a genuine gain from planning rather than evaluation noise.

The per-scenario breakdown reveals a complementary pattern. On the simpler scenarios (\scn{Straight} and \scn{Merge}), SAC and SAC + MPPI achieve the highest success rates, consistent with their tendency to overfit to easier modes of the scenario distribution. However, these gains come at the cost of substantially harsher control than \method{}. Moreover, they collapse on the more challenging scenarios (\scn{Intersection} and \scn{Roundabout}), where success rates drop below 25\%, while \method{} maintains success rates above 50\%. 

Table~\ref{tab:planner_multi} shows that this advantage persists across terminal-cost settings and how it develops as the high-level policy is trained. When the high-level policy is only partially trained (3.5M steps), \method{} is competitive but not yet uniformly dominant, suggesting that MPPI cannot fully compensate for an immature prior. As the prior improves, however, the benefits of the hierarchical decomposition become more pronounced. At 7M steps, \method{} clearly outperforms all baselines and remains robust to cost tuning.

\begin{figure}[t]
    \centering
    \includegraphics[width=0.7\linewidth]{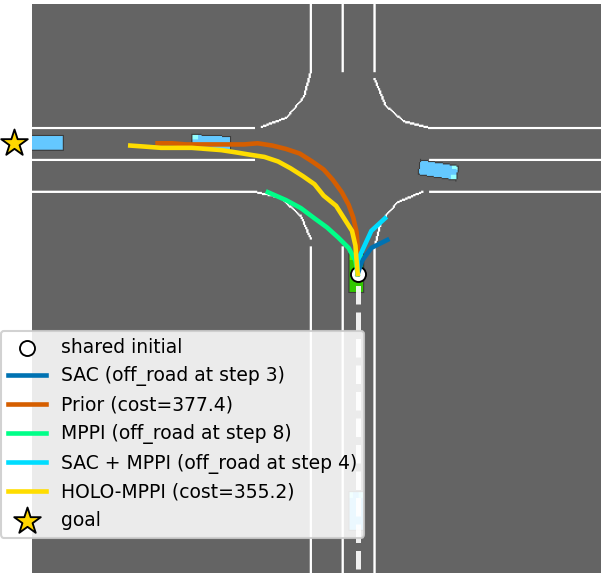}
    \caption{20-step (10 s) trajectory comparison in an intersection scenario. Starting from the same initial state, vanilla MPPI and SAC-based variants select poor turning modes and go off-road, whereas the learned high-level policy indentifies the correct maneuver. \method{} further refines this high-level intention via online MPPI, yielding a lower-cost trajectory that reaches the goal more efficiently while executing a smoother, lower-curvature turn.}
    \label{fig:trajectories}
\end{figure}

\begin{figure}[t]
    \centering
    \includegraphics[width=1.\linewidth]{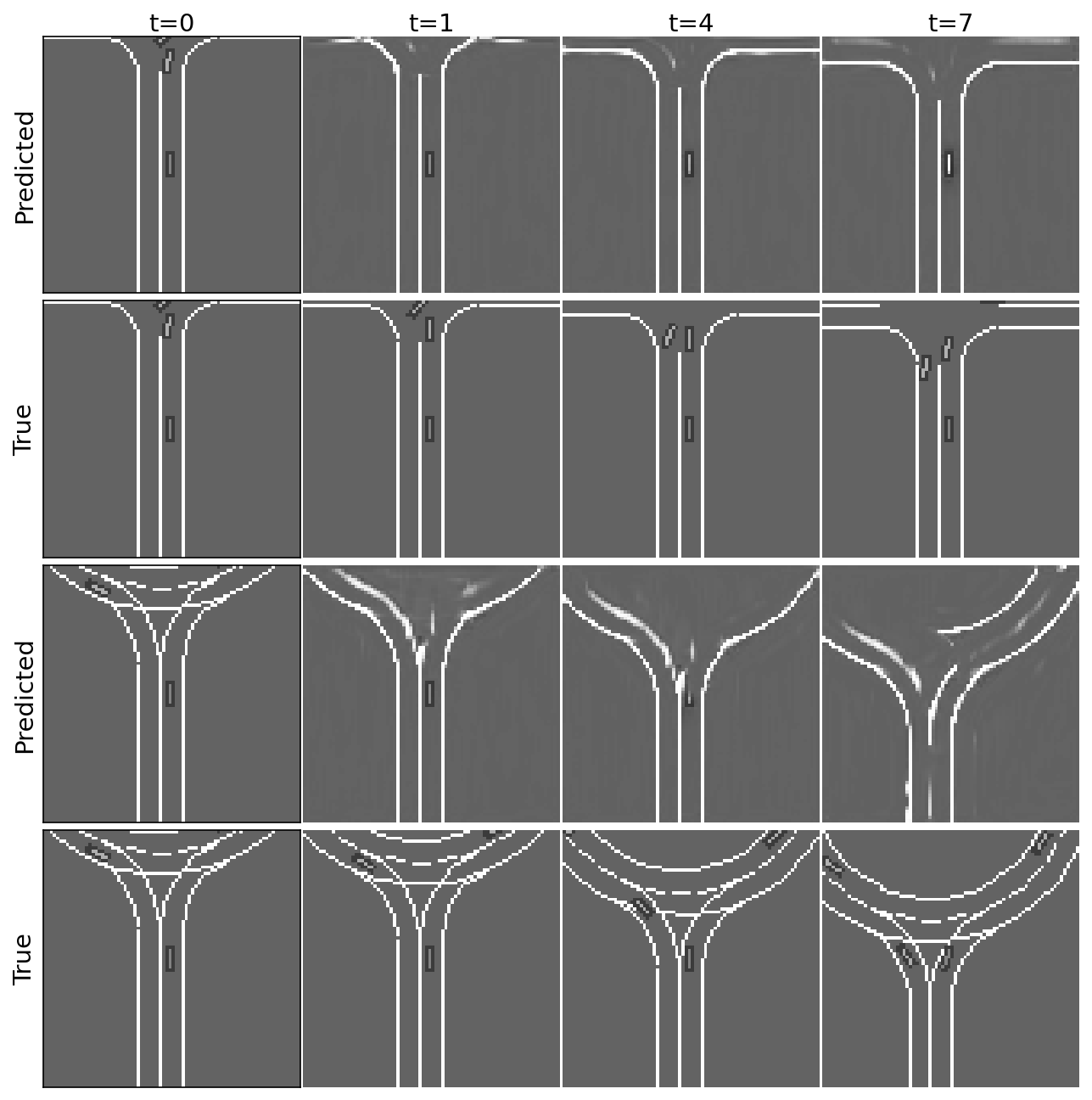}
    \caption{Qualitative comparison of the learned world model over a rollout horizon in \scn{Intersection} and \scn{Roundabout} scenarios. For each scenario, the top row shows the evolution of the predicted BEV images, and the bottom row shows the corresponding ground-truth rollout. The model preserves coarse scene structure, but visible drift accumulates over longer horizons.}
    \label{fig:world}
\end{figure}

The qualitative results are consistent with these quantitative trends. In Figure~\ref{fig:trajectories}, vanilla MPPI and the SAC-based variants select poor turning modes in the intersection example and go off-road, whereas the learned high-level policy captures the correct maneuver. \method{} then refines that intention into a trajectory that stays closer to the lane geometry and reaches the goal more cleanly than the prior alone. Figure~\ref{fig:world} further shows that the learned world model preserves the coarse scene structure well enough to support rollout-based evaluation, although visible drift remains over longer rollouts.

Finally, Table~\ref{tab:plan_time} shows that
% these gains do not come at the cost of impractical runtime. \method{} incurs additional computation relative to vanilla MPPI because it must generate a learned prior and convert the high-level action into a low-level control sequence, but the runtime remains compatible with real-time 5Hz control.
the additional computation for prior generation and action conversion still leaves runtime compatible with 5Hz real-time control.

\begin{table}[t]
    \centering
    \caption{Planning time comparison. All methods are real-time capable with 5Hz operation.}
    \setlength{\tabcolsep}{2pt}
    \begin{tabular}{@{}c|ccccc}
        \toprule
        \textbf{Method} & SAC & Prior & MPPI & SAC \!+\! MPPI & \method{} \\
        \hline\rule{0pt}{2.5ex}
        \textbf{Time (ms)} & $1.96 \pmstd{0.29}$ & $1.64 \pmstd{0.18}$ & $130 \pmstd{16}$ & $157 \pmstd{16}$ & $169 \pmstd{28}$ \\
        \bottomrule
    \end{tabular}
    \label{tab:plan_time}
\end{table}

\section{Conclusion}
We presented \method{}, a hierarchical framework for multi-scenario motion planning that pairs a high-level learned prior with low-level online MPPI refinement.
By learning in an abstract action space and refining in the original control space at test time, \method{} achieves robust performance across diverse scenarios without per-scenario retuning. In our autonomous-driving experiments, this combination improves the success rate over vanilla MPPI and end-to-end RL baselines while maintaining smooth, real-time control.

Several directions remain for future work. Extending the world model to learn neighbor behavior could improve rollouts in interactive, less structured environments such as urban driving. Scaling to a broader set of scenarios and evaluating on held-out scenario types would further test the generalization of the learned prior. End-to-end training with differentiable MPPI in the loop could align the high-level policy more directly with downstream refinement. Finally, applying \method{} to other robotic domains would test the generality of the hierarchical design.

%%%%%%%%%%%%%%%%%%%%%%%%%%%%%%%%%%%%%%%%%%%%%%%%%%%%%%%%%%%%%%%%%%%%%%%%%%%%%%%%

\bibliographystyle{IEEEtran}
\bibliography{refs}

@inproceedings{cobbe2019quantifying,
  title={Quantifying generalization in reinforcement learning},
  author={Cobbe, Karl and Klimov, Oleg and Hesse, Chris and Kim, Taehoon and Schulman, John},
  booktitle={International conference on machine learning},
  pages={1282--1289},
  year={2019},
  organization={PMLR}
}

@article{hadfield2017inverse,
  title={Inverse reward design},
  author={Hadfield-Menell, Dylan and Milli, Smitha and Abbeel, Pieter and Russell, Stuart J and Dragan, Anca},
  journal={Advances in neural information processing systems},
  volume={30},
  year={2017}
}

@article{dulac2021challenges,
  title={Challenges of real-world reinforcement learning: definitions, benchmarks and analysis},
  author={Dulac-Arnold, Gabriel and Levine, Nir and Mankowitz, Daniel J and Li, Jerry and Paduraru, Cosmin and Gowal, Sven and Hester, Todd},
  journal={Machine Learning},
  volume={110},
  number={9},
  pages={2419--2468},
  year={2021},
  publisher={Springer}
}

@article{williams2018information,
  title={Information-theoretic model predictive control: Theory and applications to autonomous driving},
  author={Williams, Grady and Drews, Paul and Goldfain, Brian and Rehg, James M and Theodorou, Evangelos A},
  journal={IEEE Transactions on Robotics},
  volume={34},
  number={6},
  pages={1603--1622},
  year={2018},
  publisher={IEEE}
}

@article{rubinstein1999cross,
  title={The cross-entropy method for combinatorial and continuous optimization},
  author={Rubinstein, Reuven},
  journal={Methodology and computing in applied probability},
  volume={1},
  number={2},
  pages={127--190},
  year={1999},
  publisher={Springer}
}

@inproceedings{pinneri2021sample,
  title={Sample-efficient cross-entropy method for real-time planning},
  author={Pinneri, Cristina and Sawant, Shambhuraj and Blaes, Sebastian and Achterhold, Jan and Stueckler, Joerg and Rolinek, Michal and Martius, Georg},
  booktitle={Conference on Robot Learning},
  pages={1049--1065},
  year={2021},
  organization={PMLR}
}

@inproceedings{yin2022trajectory,
  title={Trajectory distribution control for model predictive path integral control using covariance steering},
  author={Yin, Ji and Zhang, Zhiyuan and Theodorou, Evangelos and Tsiotras, Panagiotis},
  booktitle={2022 International Conference on Robotics and Automation (ICRA)},
  pages={1478--1484},
  year={2022},
  organization={IEEE}
}

@article{mohamed2025toward,
  title={Toward efficient MPPI trajectory generation with unscented guidance: U-MPPI control strategy},
  author={Mohamed, Ihab S and Xu, Junhong and Sukhatme, Gaurav S and Liu, Lantao},
  journal={IEEE Transactions on Robotics},
  volume={41},
  pages={1172--1192},
  year={2025},
  publisher={IEEE}
}

@inproceedings{pravitra2020L,
  title={$\mathcal{L}_1$-adaptive mppi architecture for robust and agile control of multirotors},
  author={Pravitra, Jintasit and Ackerman, Kasey A and Cao, Chengyu and Hovakimyan, Naira and Theodorou, Evangelos A},
  booktitle={2020 IEEE/RSJ International Conference on Intelligent Robots and Systems (IROS)},
  pages={7661--7666},
  year={2020},
  organization={IEEE}
}

@article{ryu2025iann,
  title={IANN-MPPI: Interaction-Aware Neural Network-Enhanced Model Predictive Path Integral Approach for Autonomous Driving},
  author={Ryu, Kanghyun and Sung, Minjun and Gupta, Piyush and D'sa, Jovin and Tariq, Faizan M and Isele, David and Bae, Sangjae},
  journal={arXiv preprint arXiv:2507.11940},
  year={2025}
}

@inproceedings{levine2013guided,
  title={Guided policy search},
  author={Levine, Sergey and Koltun, Vladlen},
  booktitle={International conference on machine learning},
  pages={1--9},
  year={2013},
  organization={PMLR}
}

@article{song2022policy,
  title={Policy search for model predictive control with application to agile drone flight},
  author={Song, Yunlong and Scaramuzza, Davide},
  journal={IEEE Transactions on Robotics},
  volume={38},
  number={4},
  pages={2114--2130},
  year={2022},
  publisher={IEEE}
}

@article{celestini2024transformer,
  title={Transformer-based model predictive control: Trajectory optimization via sequence modeling},
  author={Celestini, Davide and Gammelli, Daniele and Guffanti, Tommaso and D'Amico, Simone and Capello, Elisa and Pavone, Marco},
  journal={IEEE Robotics and Automation Letters},
  volume={9},
  number={11},
  pages={9820--9827},
  year={2024},
  publisher={IEEE}
}

@inproceedings{hansen2022temporal,
  title={Temporal Difference Learning for Model Predictive Control},
  author={Hansen, Nicklas A and Su, Hao and Wang, Xiaolong},
  booktitle={International Conference on Machine Learning},
  pages={8387--8406},
  year={2022},
  organization={PMLR}
}

@article{qu2023rl,
  title={RL-driven MPPI: Accelerating online control laws calculation with offline policy},
  author={Qu, Yue and Chu, Hongqing and Gao, Shuhua and Guan, Jun and Yan, Haoqi and Xiao, Liming and Li, Shengbo Eben and Duan, Jingliang},
  journal={IEEE Transactions on Intelligent Vehicles},
  volume={9},
  number={2},
  pages={3605--3616},
  year={2023},
  publisher={IEEE}
}

@article{jenelten2024dtc,
  title={Dtc: Deep tracking control},
  author={Jenelten, Fabian and He, Junzhe and Farshidian, Farbod and Hutter, Marco},
  journal={Science Robotics},
  volume={9},
  number={86},
  pages={eadh5401},
  year={2024},
  publisher={American Association for the Advancement of Science}
}

@article{zhang2026sumo,
  title={Sumo: Dynamic and Generalizable Whole-Body Loco-Manipulation},
  author={Zhang, John Z and Sorokin, Maks and Br{\"u}digam, Jan and Hung, Brandon and Phillips, Stephen and Yershov, Dmitry and Niroui, Farzad and Zhao, Tong and Fermoselle, Leonor and Zhu, Xinghao and others},
  journal={arXiv preprint arXiv:2604.08508},
  year={2026}
}

@inproceedings{yang2023cajun,
  title={Cajun: Continuous adaptive jumping using a learned centroidal controller},
  author={Yang, Yuxiang and Shi, Guanya and Meng, Xiangyun and Yu, Wenhao and Zhang, Tingnan and Tan, Jie and Boots, Byron},
  booktitle={Conference on Robot Learning},
  pages={2791--2806},
  year={2023},
  organization={PMLR}
}

@article{cheng2025rambo,
  title={RAMBO: RL-augmented Model-based Whole-body Control for Loco-manipulation},
  author={Cheng, Jin and Kang, Dongho and Fadini, Gabriele and Shi, Guanya and Coros, Stelian},
  journal={IEEE Robotics and Automation Letters},
  year={2025},
  publisher={IEEE}
}

@article{sutton1999between,
  title={Between MDPs and semi-MDPs: A framework for temporal abstraction in reinforcement learning},
  author={Sutton, Richard S and Precup, Doina and Singh, Satinder},
  journal={Artificial intelligence},
  volume={112},
  number={1-2},
  pages={181--211},
  year={1999},
  publisher={Elsevier}
}

@inproceedings{vezhnevets2017feudal,
  title={Feudal networks for hierarchical reinforcement learning},
  author={Vezhnevets, Alexander Sasha and Osindero, Simon and Schaul, Tom and Heess, Nicolas and Jaderberg, Max and Silver, David and Kavukcuoglu, Koray},
  booktitle={International conference on machine learning},
  pages={3540--3549},
  year={2017},
  organization={PMLR}
}

@article{nachum2018data,
  title={Data-efficient hierarchical reinforcement learning},
  author={Nachum, Ofir and Gu, Shixiang Shane and Lee, Honglak and Levine, Sergey},
  journal={Advances in neural information processing systems},
  volume={31},
  year={2018}
}

@article{schakkal2025hierarchical,
  title={Hierarchical vision-language planning for multi-step humanoid manipulation},
  author={Schakkal, Andr{\'e} and Zandonati, Ben and Yang, Zhutian and Azizan, Navid},
  journal={arXiv preprint arXiv:2506.22827},
  year={2025}
}

@misc{highway-env,
  author = {Leurent, Edouard},
  title = {An Environment for Autonomous Driving Decision-Making},
  year = {2018},
  publisher = {GitHub},
  journal = {GitHub repository},
  howpublished = {\url{https://github.com/eleurent/highway-env}},
}

@article{leurent2019social,
  title={Social attention for autonomous decision-making in dense traffic},
  author={Leurent, Edouard and Mercat, Jean},
  journal={arXiv preprint arXiv:1911.12250},
  year={2019}
}

@inproceedings{gottschalk1996obbtree,
  title={OBBTree: A hierarchical structure for rapid interference detection},
  author={Gottschalk, Stefan and Lin, Ming C and Manocha, Dinesh},
  booktitle={Proceedings of the 23rd annual conference on Computer graphics and interactive techniques},
  pages={171--180},
  year={1996}
}

%%%%%%%%%%%%%%%%%%%%%%%%%%%%%%%%%%%%%%%%%%%%%%%%%%%%%%%%%%%%%%%%%%%%%%%%%%%%%%%%

%%%%%%%%%%%%%%%%%%%%%%%%%%%%%%%%%%%%%%%%%%%%%%%%%%%%%%%%%%%%%%%%%%%%%%%%%%%%%%%%
% \onecolumn
% \section{APPENDIX}

%%%%%%%%%%%%%%%%%%%%%%%%%%%%%%%%%%%%%%%%%%%%%%%%%%%%%%%%%%%%%%%%%%%%%%%%%%%%%%%%

% \section*{Acknowledgements}

% We thank people for discussions.

\end{document}